# DEEP LEARNING BASED MOOD TAGGING FOR CHINESE SONG LYRICS

Jie Wang[1,2], Yilin Yang[3]

[1]Department of Electrical Engineering, The Hong Kong Polytechnic University
[2]Tencent Group
[3]School of Cyberspace Security, Beijing University of Posts and Telecommunications

jayjay.wang@connect.polyu.hk, yyllh7@bupt.edu.cndu.cn

## ABSTRACT

Nowadays, listening music has been and will always be an indispensable part of our daily life. In recent years, sentiment analysis of music has been widely used in the information retrieval systems, personalized recommendation systems and so on. Due to the development of deep learning, this paper commits to find an effective approach for mood tagging of Chinese song lyrics. To achieve this goal, both machine-learning and deep-learning models have been studied and compared. Eventually, a CNN-based model with pre-trained word embedding has been demonstrated to effectively extract the distribution of emotional features of Chinese lyrics, with at least 15 percentage points higher than traditional machine-learning methods (i.e. TF-IDF+SVM and LIWC+SVM), and 7 percentage points higher than other deep-learning models (i.e. RNN, LSTM). In this paper, more than 160,000 lyrics corpus has been leveraged for pre-training word embedding for mood tagging boost.

## 1. INTRODUCTION

In recent years, users of mobile music reached 512 million, constituting 68.0% of total mobile netizens and recording an annual increase of 43.81 million [1]. As a result, music has been an essential part of people's daily lives. With the rapid development of the Internet, the number and categories of on-line songs have rapidly accumulated, requiring an effective and computing-efficient way to manage huge music streams. As an important communication medium, music carries rich information, including sentiment, the essential characteristic and connotation. For this reason, sentiment analysis of music has been widely exploited in the music information retrieval and recommendation systems, and even musical therapy.

Conventional musical sentiment analysis mostly focuses on audio medium, which has a history of nearly 20 years. However, it's still difficult to extract audio emotional features accurately, and failed to reach a satisfactory level for online deployment [2]. Except for music audio, lyrics even contain rich information, including story backgrounds and emotional feelings. Moreover, lyrics could be easily processed based on text analysis methods and compatibly deployed in current systems. Thus, musical sentiment analysis will be conducted based on lyrics. Leveraged with enormous Chinese lyrics, a deep-learning based model will be trained for mood tagging into four tags, i.e. happiness, catharsis, sadness, and quiet. Vector representations of each Chinese word will pre-trained with a Word2Vector model, and then fed as an input into a convolutional neural network(CNN). In order to demonstrate the rationality and applicability of the model, several models including SVM and RNN have been studied and compared.

The rest of this paper is organized as follows: Section 2 introduces related work on the sentiment analysis of music. Section 3 describes main models used in the paper, including SVM, RNN and CNN. Experimental settings and results are discussed in Section 4. Finally, Section 5 summarizes this paper and potential future work.

## 2. RELATED WORK

Music emotion classification has received intense attention in the academic field. Most of previous studies concentrate on audio analysis [3], [4]. Moreover, multi-model approaches have been proposed based on both audio and lyrics [5], [6]. From the perspective of text-oriented sentiment classification research, prior reports mostly focus on emotional polarity analysis, which divides the sentiment of lyrics into positive and negative. To address this issue, there are two main ways: building emotional dictionaries [7], [8], and using machine learning [9-11]. The advantages of using sentiment lexicon and rules for text sentiment analysis are its finer granularity and capability of explicit analysis. However, this solution cannot capture significant patterns hidden in data and also require intense manpower for expert rules. The sentiment analysis based on machine learning overcomes the influence of sparse emotional words, however, it needs more elaborate feature engineering.

With the rapid development of deep learning, the state of the art results on sentiment analysis have been reported [12], [13]. However, there are still few works on sentiment analysis and mood tagging of music lyrics, which is very valuable for analysis of users' portraits on social media. The concept of deep learning was originally proposed by Hinton et al. [15], [16] in 2006. It simulates the layered processing mechanism of human brain for visual signals and can extract hierarchical features from complex raw data. The features extracted by deep learning can be regarded as the abstract representation of the original data at a higher level, which is very suitable for solving some relatively abstract recognition tasks. Since its birth, deep learning has achieved many excellent research results in many fields such as computer vision [17], speech recognition [18] and natural language processing [19,20].

## 3. MODEL

### 3.1 Machine Learning

Support Vector Machine (SVM) is an extensively exploited classification model, which has been widely demonstrated with better performance than other machine learning methods [21]. The reason SVM is so powerful is that it can transform input low-dimensional spatial data into high-dimensional space through the kernel function [22]. A segmentation plane in the high-dimensional kernel space should be found to make the interval between the training points on both sides largest. The kernel function of SVM applied in this paper is Radial Basis Function (RBF). In addition to the mainframe model, feature selection is also important for mode building. Here, two kinds of lexicon features have been used and compared, i.e. *tf-idf* and LIWC.

#### 3.1.1 TF-IDF

The main idea of the term frequency with inverse document frequency (*tf-idf*) is defined as follows: *tf* is the time of a word appears in a document which means that higher the frequency of a word occurs in a specific document, the more important it is for the document distinguishing; *idf* is the quantification of the word distributions in all documents, meaning that more times a word appears in all documents, the less likely it is attributed to documents classification [23]. Thus, the *ti-idf* term can be calculated as:

$$tf - idf(t,d) = f_{(t,d)} \log |D|/f_t \qquad (1)$$

#### 3.1.2 LIWC

LIWC (Linguistic Inquiry and Word Count) is a kind of NLP technology for word analysis and classification, including computing the percentage of causal words, emotional words, cognitive words and other psychological words in the entire text. LIWC consists of 4 general descriptive categories (total number of words, number of words per sentence, number of words that has more than six-letter, crawl rate), 22 language feature categories (i.e. personal pronouns, auxiliary verbs, conjunctions, prepositions), 32 categories of psychological characteristics (i.e. social process words, emotional process words, cognitive process words, physiological process words, etc.), 7 personalized categories (i.e. work, leisure, family, money, etc.), 3 paralinguistic categories (such as due words, pause words, fillers, etc.) and 12 punctuation categories (such as periods, commas, colons, semicolons, etc.). There are more than 80 word categories and about 4,500 words [24]. The dictionary can be defined as a collection of lexical items: $L = \{(W, C, X)\}$, where $W$ stands for words, $C$ represents the category of word and $X$ donates the weight of $W$ (if $X$ is 1, there is no weight).

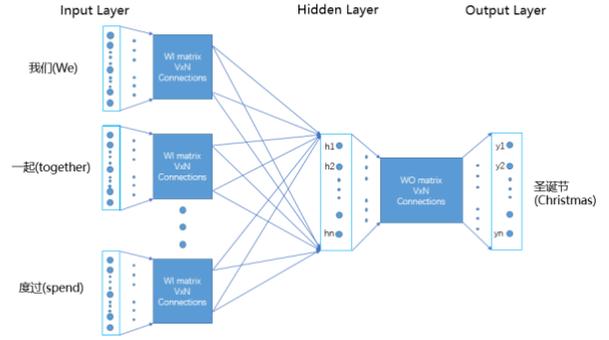

**Figure 1.** The network structure of CBOW model.

### 3.1 Deep Learning

#### 3.1.1 Word2Vector

Word embedding projecting words into a dense-dimensional vector space is essentially a feature learner which encodes word-level semantic features [14]. There exist several ways to produce a distributed representation of words, typically including continuous bag-of-words (CBOW) and Skip-gram. As shown in **Figure 1**, CBOW predicts current word based on surrounding context words while Skip-gram uses current word to predict surrounding context words. In this paper, we use the CBOW model to train a pre-trained word vector.

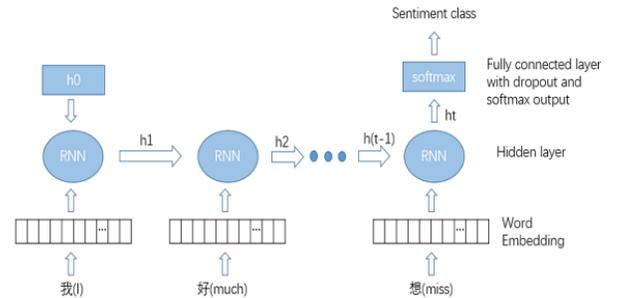

**Figure 2.** The network structure of RNN model.

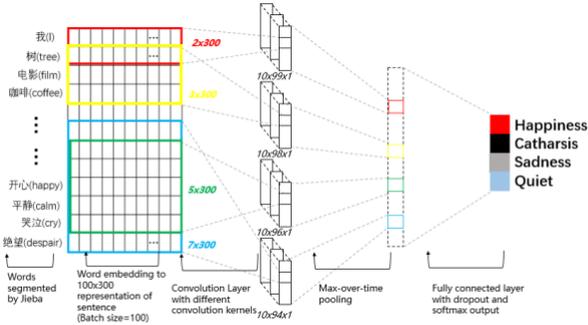

**Figure 3.** The architecture of CNN model.

*3.1.2 Recurrent Neural Network*

Recurrent Neural Network (RNN) differs from Dense Neural Network (DNN) or Convolution Neural Network (CNN) in terms of propagation of historical information via a chain-like network architecture, as depicted in **Figure 2**. Cyclic neural networks can be extended for longer sequence data, where it looks at the current input $x_t$ as well as previous hidden state $h_{t-1}$ at each time step.

However, standard RNN suffers from gradient vanishing especially when input sequence is very long. To address this issue, LSTM (Long Stort Term Memory Network) was first proposed by Hochreiter and Schmidhuber [25]. As a variant of RNN, LSTM is capable to learn long-term dependence of sequence.

*3.1.3 Convolution Neural Network*

CNN has been widely used as the core deep-learning unit of computer vision. As shown in **Figure 3**, the bottom layer is composed of convolutional layer and pooling layer. The top layer generally uses the full connection layer for specific tasks. Because of its excellent performance and great success in image recognition tasks, CNN has been extended into more research fields by researchers [17]. This paper also attempts to apply CNN for mood tagging of lyrics. Such a special network structure makes CNN own the following significant advantages:

- The alternating superposition of convolutional layer and pooling layer promotes the representation capability of local features.
- Combination of feature extraction and classification task enables end-to-end model training.
- Local sensation and weight sharing of convolution layer facilitates parallel computing and make model train and inference very fast with help of GPU.

As shown in **Figure 3**, the overall CNN consists of 4 layers, i.e. a 1D-convolutional layer with a tanh activation, a batch-normalization layer, a max-pooling layer, and a flatten layer. The convolutional layer uses 4 different kernel sizes to extract multiple local feature maps. After normalization, the most representative features of each feature map are extracted by the max-pooling layer. and then go through the flatten layer. In the train phase, drop out with rate of 0.5 is used. At last, classification of 4 categories is completed through a full-connected layer with a softmax function.

## 4. EXPERIMENTS

### 4.1 Data Preprocessing

On the premise of balanced samples, the existing songs with sentiment labels are divided into 4 categories: happiness, catharsis, sadness, and quiet. Then, according to song name and singer, we use the reptile technique to obtain the vital parameter of the song: music ID. In this way we can download the corresponding lyrics from QQ music. This process is shown in Figure 3.

After de-duplication and screening, the total number of effective lyrics with sentiment labels is 11,427, among which there are 2,870 happy songs, 2,812 angry songs, 2,848 sad songs, and 2,897 quiet songs.

After the dataset is divided, we need to perform a series of preprocessing (as shown in Figure 4):
- Filter out time axis, song introduction, punctuation and special characters in the lyric files, and only keep Chinese text with more semantic information.
- Word segmentation. Different from English, there is no way to distinguish between words and words in Chinese text. If want to modeling, we need to first process word segmentation. The final experimental effect is related to the quality of word segmentation. In this processing, we use the jieba tool.
- Training word embedding. In this paper, word embedding is trained using CBOW model, which has been implemented by the Word2Vec tool and can be used directly. It is worth mentioning that in order to get higher quality of word embedding, the corpus used for training should not be too small. Therefore, we crawled 160,000 lyrics as input for training. And then we can obtain the 300-dimensional vector of each word.

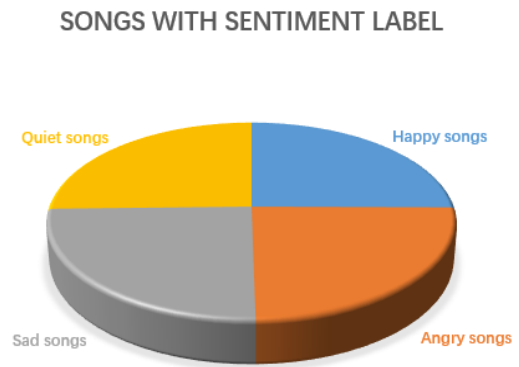

**Figure 4.** Distribution of four kinds of lyrics.

**Table 1.** Results of comparative experiment.

| Models | Machine Learning | | Deep Learning | | |
|---|---|---|---|---|---|
| | TF-IDF + SVM | LIWC + SVM | CNN | RNN | LSTM |
| Accuracy (%) | 47.059 | 55.35 | 72.73 | 64.13 | 65.617 |

### 4.2 Model Implementation

Of all samples, the training sets to the test sets ratio is 9 to 1. While training, we only take the first 100 participles of each song as input for word2vec, and then we can get the corresponding 100x300 feature matrix, which is transmitted as the input to the convolutional neural networks.

At present, the most commonly used training methods for convolutional neural networks are the traditional gradient descent. Among them, the Batch Gradient Descent method could find the optimal solution finally, but since all the samples are needed to participate in the operation in each weight update, the convergence speed is much too slow. the Stochastic Gradient Descent method only needs one sample to participate in the operation each time when the weight is updated. Therefore, the convergence speed can be significantly accelerated, but it is so easy to cause convergence to the local optimal solution. In order to give consideration to the advantages of the two methods, this paper uses the Mini-batch Gradient Descent method for training, that is, each weight update requires a small number of samples to participate in the calculation, which can improve the training speed and, at the same time, ensure that the optimal solution is found as much as possible. There are 11,427 samples in this dataset. Through trial and error, a good compromise can be obtained when the size of batch is set to 100.

In order to prevent over-fitting, the L2 regularization is used to constrain the network's parameters. And the dropout strategy is introduced in the training process of the fully connected layer. We make dropout equal to 0.5 during the training process, that is, half of the parameters were randomly discarded. What's more, the optimizers we used for training is Adam, which works better than SGD proved by experiment.

**Table 2.** Evaluation indexes of CNN.

| | Precision | Recall | F1-score | Support |
|---|---|---|---|---|
| Happiness | 0.81 | 0.73 | 0.77 | 295 |
| Catharsis | 0.69 | 0.77 | 0.72 | 265 |
| Sadness | 0.72 | 0.82 | 0.77 | 274 |
| Quiet | 0.69 | 0.63 | 0.66 | 309 |
| Avg/Total | 0.73 | 0.73 | 0.73 | 1143 |

**Table 3.** LIWC frequency distribution of different songs.

| Sad Song | Quiet Song |
|---|---|
| 爱(love) | 爱(love) |
| 说(say) | 说(say) |
| 想(think) | 想(think) |
| 走(walk) | 里(inside) |
| 爱情(love) | 我会(I will) |
| 回忆(memory) | 走(walk) |
| 心(heart) | 永远(forever) |
| 里(inside) | 世界(world) |
| 寂寞(loneliness) | 啊(ah) |
| 幸福(happiness) | 心(heart) |
| 快乐(happy) | 中(in) |
| 离开(leave) | 感觉(feel) |
| 时间(time) | 请(please) |
| 世界(World) | 告诉(tell) |
| 哭(cry) | 做(do) |
| 太(too) | 离开(leave) |
| 做(do) | 时间(time) |
| 永远(forever) | 爱情(love) |
| 我会(I will) | 想要(want) |
| 中(among) | 宝贝(baby) |
| 眼泪(tear) | 希望(hope) |
| 真的(really) | 太(too) |
| 懂(kown) | 天空(sky) |
| 听(listen) | 身边(side) |
| 忘记(forget) | 真的(really) |
| 笑(smell) | 听(listen) |
| 忘(forget) | 回忆(memory) |
| 记得(remember) | 生活(life) |
| 梦(dream) | 梦(dream) |

### 4.3 Result

In order to demonstrate the feasibility and applicability of CNN in sentiment classification of Chinese lyrics, we designed a series of comparative experiments. The experimental results are shown in Table 1.

- Compared with the traditional machine learning method, the CNN model proposed in this paper has achieved better performance in the sentiment classification task, and eliminates the complicated steps of manually extracting features. The accuracy of CNN is more than 15% higher than SVM, which also proves the robustness of the CNN structure in the noisy data environment.

- It has been proved that the CNN method which extracts multiple local features by designing different convolution kernels is better than the RNN or LSTM model with basic structure. In other words, the CNN model is more suitable for emotional detection tasks.
- In addition, we have also made a comparison experiment with the corpus. The classification accuracy of using 160,000 lyrics as corpus is 2% higher than that of 2G-sized Baidu Encyclopedia corpus. This also proves that the purity of corpus is more important than the size of corpus.
- As can be seen from Table 2, quiet songs are more easily misclassified into other classes. This is mainly because the emotional expression of quiet songs is not as prominent as other categories, and there are not many representative words and explicit emotional features. This is also part of the reason for the low accuracy of SVM method in this sentiment classification task.

Another reason is that the word frequency distribution of quiet songs overlaps with other classes'. The word frequency distribution of quiet songs and sad songs are shown in Table 3.

## 5. CONCLUSION AND FUTURE WORK

Sentiment classification has always been one of the important tasks in NLP. This paper attempts to solve the problem of sentiment classification of Chinese lyrics by using the idea of deep learning. First of all, this paper uses Word2Vec model to automatically train the vector representation of all the words in the 160,000 lyrics, and then combines them into a two-dimensional matrix as the emotional features of each sample. In this way, the cumbersome steps of manually extracting features are eliminated. Secondly, this paper proposes a convolutional network structure with 4 different convolution kernels, so that the feature representation of various granularity can be obtained. The comparative experiments with the traditional machine learning models show that CNN shows much better performance. In addition, through comparison experiments with other deep learning models, it is proved that this structure can better describe the sentiment feature distribution of Chinese lyrics. In the future, we will continue to optimize the network structure to achieve higher accuracy, and will continue to analyze the deep reasons of classification errors. In addition, we will consider combining lyrics with audio files and using deep learning method to predicting sentiment tendencies in the songs.